\documentclass{llncs}

\usepackage{latexsym}
\usepackage{float}
\usepackage{amssymb}
\usepackage{amsmath}
\usepackage{url}

\floatstyle{ruled}
\newfloat{algorithm}{tbp}{loa}
\floatname{algorithm}{Figure}    

\newcommand{\cpilebs}{\mathit{compile\mbox{-}basic}}
\newcommand{\cpilebc}{\mathit{compile\mbox{-}bc}}
\newcommand{\cpileuc}{\mathit{compile\mbox{-}uc}}
\newcommand{\bc}{\mathit{break\mbox{-}count}}

\newcommand{\ls}{\mathit{ls}}
\newcommand{\flp}{\mathit{Flip}}
\newcommand{\plc}{\mathit{PL}^{cc}}
\newcommand{\wcc}{\mathit{WSAT}^{cc}}
\newcommand{\wsat}{\mathit{WSAT}}
\newcommand{\zch}{\mathit{zchaff}}
\newcommand{\vwcc}{\mathit{vb\mbox{-}WSAT}^{cc}}
\newcommand{\gwcc}{\mathit{Generic\mbox{-}WSAT}^{cc}}
\newcommand{\dwcc}{\mathit{df\mbox{-}WSAT}^{cc}}
\newcommand{\col}{\mathit{col}}
\newcommand{\vc}{\mathit{vc}}
\newcommand{\mtr}{\mathit{Max\mbox{-}Tries}}
\newcommand{\mfl}{\mathit{Max\mbox{-}Flips}}

\newcommand{\At}{\mathit{At}}
\newcommand{\size}{\mathit{size}}

\newcommand{\tr}{\mathbf{t}}
\newcommand{\fa}{\mathbf{f}}

\title{Local-search techniques for propositional logic extended
with cardinality constraints} 

\titlerunning{Local-search techniques for propositional logic with
cardinality constraints}

\author{Lengning Liu \and Miros\l aw Truszczy\'nski}
\authorrunning{Lengning Liu and Miros\l aw Truszczy\'nski}
\institute{Department of Computer Science, University
of Kentucky, Lexington, KY 40506-0046, USA}

\begin{document}

\maketitle
\vspace*{-0.06in}
\begin{abstract}
We study local-search satisfiability solvers for propositional logic 
extended with cardinality atoms, that is, expressions that provide
explicit ways to model constraints on cardinalities of sets. Adding 
cardinality atoms to the language of propositional logic facilitates 
modeling search problems and often results in concise encodings. 
We propose two ``native'' local-search solvers for theories in the 
extended language. We also describe techniques to reduce the problem 
to standard propositional satisfiability and allow us to use 
off-the-shelf SAT solvers. We study these methods experimentally.
Our general finding is that native solvers designed specifically for 
the extended language perform better than indirect methods relying on 
SAT solvers.
\end{abstract}

\section{Introduction}
\label{intro}

We propose and study local-search satisfiability solvers for an 
extension of propositional logic with explicit means to represent
cardinality constraints.

In recent years, propositional logic has been attracting considerable
attention as a general-purpose modeling and computing tool, well 
suited for solving search problems. For instance, to solve a graph 
$k$-coloring problem for an undirected graph $G$, we construct a 
propositional theory $T$ so that its models encode $k$-colorings of 
$G$ and there is a polynomial-time method to reconstruct a 
$k$-colorings of $G$ from a model of $T$. Once we have such a theory 
$T$, we apply to it a satisfiability solver, find a model of $T$ 
and reconstruct from the model the corresponding 
$k$-coloring of $G$. 

Instances of many other search problems can be represented in a similar 
way as propositional 
theories and this modeling capability of the propositional logic has 
been known for a long time. However, it has been only recently that 
we saw a dramatic improvement in the performance of programs to compute 
models of propositional theories \cite{skc94,la97,zh97,ss99,mmz01,gn02}.
These new programs can often handle theories consisting of hundreds of 
thousands, sometimes millions, of clauses. They demonstrate that 
propositional logic is not only a tool to represent problems but also 
a viable computational formalism.

The approach we outlined above has its limitations. The 
repertoire of operators available for building formulas to represent
problem constraints is restricted to boolean connectives. Moreover, 
since satisfiability solvers usually require CNF theories as input, 
for the most part the only formulas one can use to express constraints 
are clauses. One effect of these restrictions is often very large 
size of CNF theories needed to represent even quite simple constraints
and, consequently, poorer effectiveness of satisfiability solvers in
computing answers to search problems. Researchers recognized this 
limitation of propositional logic. They proposed extensions to the 
basic language with the equivalence operator \cite{li00}, with
cardinality atoms \cite{bss94,et01a} and with pseudo-boolean
constraints \cite{bar95,wal97,arms02,dg02,pre02}, and developed 
solvers capable of computing models for theories in the expanded 
syntax.

In this paper, we focus on an extension of propositional logic 
with {\em cardinality atoms}, as described in \cite{et01a}.
Specifically, a cardinality atom is an expression of the form $k X 
m$, where $k$ and $m$ are non-negative integers and $X$ is a set of 
propositional atoms. Cardinality atoms offer a {\em direct} means to 
represent cardinality constraints on sets and help construct concise 
encodings of many search problems. We call this extension of the 
propositional logic the {\em propositional logic with cardinality 
constraints} and denote it by $\plc$.

To make the logic $\plc$ into a 
computational mechanism, we need programs to compute models of 
$\plc$ theories. One possible approach is to {\em compile cardinality 
atoms away}, replacing them with equivalent propositional-logic 
representations. After converting the resulting theories to CNF,
we can use any off-the-shelf satisfiability solver to compute models.
Another approach is to design solvers specifically tailored to the 
expanded syntax of the logic $\plc$. To the best of our knowledge, the
first such solver was proposed in \cite{bss94}. A more recent solver,
{\em aspps}\footnote{The acronym for answer-set programming with
propositional schemata.}, was described in \cite{et01a}. 

These two solvers are {\em complete} solvers. In this paper, we 
propose and study {\em local-search} satisfiability solvers that can
handle the extended syntax of the logic $\plc$. In our work we 
built on ideas first used in {\em WSAT}, one of the 
most effective local-search satisfiability solvers for propositional 
logic \cite{skc94}\footnote{In the paper, we write {\em WSAT} instead 
of {\em WALKSAT} to shorten the notation.}. In particular, as in 
{\em WSAT}, we proceed by executing a prespecified number of {\em 
tries}. Each try starts with a random truth assignment and consists 
of a sequence of local modification steps called {\em flips}. Each flip
is determined by an atom selected from an {\em unsatisfied} clause. 
We base the choice of an atom on the value of its {\em break-count} 
(some measure of how much the corresponding flip increases the degree 
to which the clauses in the theory are violated). In $\wsat$, the
break-count of an atom is the number of clauses that become 
unsatisfied when the truth value of the atom is flipped. In the 
presence of cardinality atoms, this simple measure does not lead to 
satisfactory algorithms and modifications are necessary.

In this paper, we propose two approaches. In the first of them, we
change the definition of the {\em break-count}. To this end, we exploit
the fact that cardinality atoms are only high-level shorthands for 
some special propositional theories and, as we already indicated
earlier, can be {\em compiled} away. Let $T$ be a $\plc$ theory and let
$T'$ be its propositional-logic equivalent. We define the break-count of 
an atom $a$ in $T$ as the number of clauses {\em in the compiled
theory} $T'$ that become unsatisfied after we flip $a$. Important thing 
to note is that we do not need to compute $T'$ explicitly in order to 
compute the break-count of $a$. It can be computed directly on 
the basis of $T$ alone.

Our second approach keeps the concept of the break-count exactly as it
is defined in {\em WSAT} but changes the notion of a flip. This
approach applies whenever a $\plc$ theory $T$ can be separated into two 
parts $T_1$ and $T_2$ so that: (1) 
$T_2$ consists of propositional clauses, (2) it is easy to construct 
random assignments that satisfy $T_1$, and (3) for every truth 
assignment satisfying $T_1$, (modified) flips executed on this 
assignment result in assignments that also satisfy $T_1$. In such 
cases, we can start a try by generating an initial truth assignment 
to satisfy all clauses in $T_1$, and then executing a sequence of
(modified) flips, choosing atoms for flipping based on the number of 
clauses in $T_2$ (which are all standard propositional CNF clauses) 
that become unsatisfiable after the flip.

In the paper, we develop and implement both ideas. We study 
experimentally the performance of our algorithms on several search 
problems: the graph coloring problem, the vertex-cover problem and 
the 
open
latin-square problem. We compare the 
performance of our algorithms to that of selected SAT solvers executed 
on CNF theories obtained from $\plc$ theories by compiling away 
cardinality atoms.  

\section{Logic $\plc$}
\label{s-plc}


The language of the logic $\plc$ is determined by the set $\At$ of 
{\em propositional atoms} and two special symbols $\bot$ and $\top$ that we
always interpret as {\em false} and {\em true}, respectively. 
A {\em cardinality atom} (c-atom, for short) is an expression of the 
form $k X m$, where $X$ is a set of propositional atoms, and $k$ and 
$m$ are non-negative integers. 
If $X=\{a_1,\ldots,a_n\}$, we will also write $k \{ a_1,\ldots, a_n\} m$ 
to denote a c-atom $k X m$. One (but not both) of $k$ and $m$ may 
be missing. Intuitively, a 
c-atom $k X m$ means: 
{\em at least $k$ and no more than $m$ of atoms in $X$
are true}. If $k$ (or $m$) is missing, the c-atom constrains the number 
of its propositional atoms that must be true only from above (only from 
below, respectively). 

A {\em clause} is an expression of the form
$
\neg \alpha_1\vee\ldots\vee \neg \alpha_r \vee \beta_1 \vee \ldots \vee 
\beta_s,
$
where each $\alpha_i$, $1\leq i\leq r$, and each $\beta_j$, $1\leq j
\leq s$, is a propositional atom or a c-atom. A {\em theory} of the 
logic $\plc$ is any set of clauses\footnote{It is easy to extend the 
language of $\plc$ and introduce arbitrary formulas built of atoms 
and c-atoms by means of logical connectives. Since clausal theories, 
as in propositional logic, are most fundamental, we focus on clausal
theories only.}. 

An {\em interpretation} is an assignment of truth values $\tr$ and
$\fa$ to atoms in $\At$.
An interpretation $I$ {\em satisfies} an atom $a$ if $I(a)=\tr$. An 
interpretation $I$ satisfies a c-atom $k\{a_1,\ldots, a_n\}m$ if 
$
k \leq |\{i\colon I(a_i)=\tr\}| \leq m.
$

This notion of satisfiability extends in a standard way to clauses
and theories. We will write interchangingly ``is a model of'' and 
``satisfies''. We will also write $I\models E$, when $I$ is a model of 
an atom, c-atom, clause or theory $E$.

We will now illustrate the use of the logic $\plc$ as a modeling tool 
by presenting $\plc$ theories that encode (1) the graph-coloring 
problem, (2) the graph vertex-cover problem, and (3) the
open
latin-square problem. We later use these theories
as benchmarks in performance tests.

In the first of these problems we are 
given a graph $G$ with the set $V=\{1,\ldots,n\}$ of vertices 
and a set $E$ of edges (unordered pairs of vertices). We 
are also given a set $C=\{1,\ldots,k\}$ of colors. The objective is
to find an assignment of colors to vertices so that for every edge, its
vertices get different colors. A $\plc$ theory representing this
problem is built of propositional atoms $c_{i,j}$, where $1\leq i\leq n$,
and $1\leq j\leq k$. An intended meaning of an atom $c_{i,j}$ is that 
{\em vertex $i$ gets color $j$}. We define the theory 
$\col(G,k)$ to consist of the following clauses:
\begin{enumerate}
\item $1\{c_{i,1},\ldots,c_{i,k}\}1$, for every $i$, $1\leq i \leq n$.
These clauses ensure that every vertex obtains {\em exactly one} color
\item $\neg c_{p,j}\vee \neg c_{r,j}$, for every edge $\{p,r\}\in E$
and for every color $j$. These clauses enforce the main colorability
constraint.
\end{enumerate}
It is easy to see that models of the theory $\col(G,k)$ are indeed in
one-to-one correspondence with $k$-colorings of $G$.

In a similar way, we construct a theory $\vc(G,k)$ that represents the 
{\em vertex-cover} problem. Let $G$ be an undirected graph with the set
$V=\{1,\ldots,n\}$ of vertices and a set $E$ of edges. Given $G$ and a 
positive integer $k$, the objective is to find a set $U$ of no more 
than $k$ vertices, such that every edge has at least one of its vertices 
in $U$ (such sets $U$ are {\em vertex covers}).
We build the theory $\vc(G,k)$ of atoms $in_i$, $1\leq i\leq
n$, (intended meaning of $in_i$: vertex $i$ is in a vertex cover) and
define it to consist of the following clauses:
\begin{enumerate}
\item $\{in_1,\ldots,in_n\}k$. This clause guarantees that at most $k$
vertices are chosen to a vertex cover
\item $in_p \vee in_r$, for every edge $\{p,r\}\in E$. These clauses 
enforce the main vertex cover constraint.
\end{enumerate}
Again, it is evident that models of theory $\vc(G,k)$ are in one-to-one
correspondence with those vertex covers of $G$ that have no more than
$k$ elements. 

In the open latin-square problem, we are given an integer $n$ and a
collection $D$ of triples $(i,j,k)$, where $i$, $j$ and $k$ are 
integers from $\{1,\ldots, n\}$. The goal is to find an $n\times n$ 
array $A$ such that all entries in $A$ are integers from $\{1,\ldots,
n\}$, no row and column of $A$ contains two identical integers, and for 
every $(i,j,k)\in D$, $A(i,j)=k$. In other words, we are looking for a
{\em latin square} of order $n$ that extends the partial assignment 
specified by $D$. To represent this problem we construct a $\plc$ theory 
$\ls(n,D)$ consisting of the following clauses:
\begin{enumerate}
\item $a_{i,j,k}$, for every $(i,j,k)\in D$ (to represent the partial 
assignment $D$ given as input)
\item $1\{a_{i,j,1},\ldots,a_{i,j,n}\}1$, for every $i,j=1,\ldots,n$
(to enforce that every entry receives exactly one value)
\item $\{a_{i,1,k},\ldots,a_{i,n,k}\}1$, for every $i,k=1,\ldots,n$
(in combination with (2) these clauses enforce that an integer $k$ 
appears exactly once in a row $i$) 
\item $\{a_{1,j,k},\ldots,a_{n,j,k}\}1$, for every $j,k=1,\ldots,n$
(in combination with (2) these clauses enforce that an integer $k$ 
appears exactly once in a column $j$). 
\end{enumerate}
One can verify that models of the theory $\ls(n,D)$ correspond to
solutions to the open latin-square problem with input $D$. 

The use of c-atoms in all these three examples results in
concise representations of the corresponding problems. Clearly, we
could eliminate c-atoms and replace the constraints they represent
by equivalent CNF theories. However, the encodings become less 
direct, less concise and more complex. 

\section{Using SAT Solvers to Compute Models of $\plc$ Theories}
\label{sat}

We will now discuss methods to find models of $\plc$ theories by
means of standard SAT solvers. A key idea is to compile away c-atoms
by replacing them with their propositional-logic descriptions. We will
propose several ways to do so.

Let us consider a c-atom $C=k\{ a_{1},...,a_{n}\} m$ and let us define
a CNF theory $C'$ to consist of the following clauses:
\begin{enumerate}
\item $\neg a_{i_{1}}\vee...\vee\neg a_{i_{m+1}}$, for any $m+1$ atoms
$a_{i_{1}},...,a_{i_{m+1}}$ from $\{ a_{1},...,a_{n}\}$ (there are $n
\choose {m+1}$ such clauses); and
\item $a_{i_{1}}\vee...\vee a_{i_{n-k+1}}$, for any $n-k+1$ atoms
$a_{i_{1}},..., a_{i_{n-k+1}}$ in $\{ a_{1},...,a_{n}\}$ (there are $n
\choose {k-1}$ such clauses).
\end{enumerate}
It is easy to see that the theory $C'$ has the same models as the
c-atom $C$.

Let $T$ be a $\plc$ theory. We denote by $\cpilebs(T)$ the CNF theory 
obtained from $T$ by replacing every c-atom $C$ with the conjunction 
of clauses in $C'$ and by applying distributivity to transform the 
resulting theory into the CNF. This approach translates $T$ into a 
theory in the same language but it is practical only if $k$ and $m$ 
are small (do not exceed, say 2). Otherwise, the size of the theory 
$\cpilebs(T)$ quickly gets too large for SAT solvers to be effective.

Our next method to compile away c-atoms depends on counting. To
simplify the presentation, we will describe it in the case of a c-atom
of the form $k X$ but it extends easily to the general case. We will 
assume that $k \geq 1$ (otherwise, $k X$ is true) and $k \leq |X|$ 
(otherwise $k X$ is false). 

Let us consider a $\plc$ theory $T$ and let us assume that $T$ contains
a c-atom of the form $C = k \{a_1,\dots,a_n\}$. We introduce new 
propositional atoms: $b_{i,j}$, $i=0,\dots,n$; $j=0,\dots,k$. The 
intended role for $b_{i,j}$ is to represent the fact that at least $j$ 
atoms in $\{a_1,\ldots,a_i\}$ are true. Therefore, we define a theory 
$C'$ to consist of the following clauses:
\begin{enumerate}
\item $b_{0,j} \leftrightarrow \bot$, $j=1,\ldots,k$,
\item $b_{i,0} \leftrightarrow \top$, $i=0,\ldots,n$,
\item $b_{i,j} \leftrightarrow b_{i-1,j} \vee(b_{i-1,j-1} \wedge
a_{i})$, $i=1,\ldots, n$, $j=1,\ldots,k$.
\end{enumerate}
Let $I$ be an interpretation such that $I\models C'$. One can verify
that $I\models b_{i,j}$ if and only if $I\models j\{a_1,\ldots, a_i\}$.
In particular, $I\models b_{n,k}$ if and only if $I\models C$. Thus,
if we replace $C$ in $T$ with $b_{n,k}$ and add to $T$ the theory $C'$
the resulting theory has the same models (modulo new atoms) as $T$. By
repeated application of this procedure, we can eliminate all c-atoms
from $T$. Moreover, if we represent theories $C'$ in CNF, the resulting
theory will itself be in CNF. We will denote this CNF theory as
$\cpileuc(T)$, where $uc$ stands for {\em unary} counting. One can
show that the size of $\cpileuc(T)$ is $O(R\times\size(T))$, where $R$ 
is the maximum of all integers appearing in $T$ as lower or upper bounds 
in c-atoms. It follows that, in general, this translation leads to more
concise theories than $\cpilebs$. However, it does introduce new
atoms.

The idea of counting can be pushed further. Namely, we can design a
more concise translation than $\cpileuc$ by following the 
idea of counting and by representing numbers in the 
binary system and by building theories to model binary counting and 
comparison. For a $\plc$ theory $T$, we denote the result of applying 
this translation method to $T$ by $\cpilebc$ ($bc$ stands for {\em
binary} counting). Due to space limitation we omit the details of this
translation. We only note that the size of $\cpilebc(T)$ is $
O(\size(T)\log_2 (R+1))$, where $R$ is the maximum of all integer
bounds of c-atoms appearing in $T$. 

\section{Local-search Algorithms for the Logic $\plc$}
\label{main}

In this section we describe a local-search algorithm $\gwcc$
designed to test satisfiability of theories in the logic $\plc$.
It follows a general pattern of {\em WSAT} \cite{skc94}. The algorithm 
executes $\mtr$ independent {\em tries}. Each try starts in a randomly 
generated truth assignment and consists of a sequence of up to $\mfl$ 
{\em flips}, that is, local changes to the current truth assignment. 
The algorithm terminates with a truth assignment that is a model of 
the input theory, or with no output at all (even though the input theory 
may in fact be satisfiable). We provide a detailed description of the 
algorithm $\gwcc$ in Figure 1.

\begin{algorithm}[t]
\caption{ Algorithm $\gwcc(T)$}
\begin{flushleft}
\ \\
{\bf INPUT}:~~~~~~$T$ - a $\plc$ theory\\
{\bf OUTPUT}:~~~$\sigma$ - a satisfying assignment of $T$, or no
output\\
{\bf BEGIN}

1. ~~~~{\bf For} $i\leftarrow1$ {\bf to} $\mtr$, {\bf do}

2. ~~~~~~~~$\sigma\leftarrow\textrm{randomly generated truth
  assignment}$;

3. ~~~~~~~~{\bf For} $j\leftarrow1$ {\bf to} $\mfl$, {\bf do}

4. ~~~~~~~~~~~~{\bf If} $\sigma\models T$ {\bf then} {\bf return} $\sigma$;

5. ~~~~~~~~~~~~$C\leftarrow$ randomly selected unsatisfied clause;

6. ~~~~~~~~~~~~{\bf For each} atom $x$ {\bf in} $C$, compute
      $\bc(x)$;

7. ~~~~~~~~~~~~{\bf If} any of these atoms has break-count 0 {\bf then}

8. ~~~~~~~~~~~~~~~~randomly choose an atom with break-count 0, call it $a$;

9. ~~~~~~~~~~~~{\bf Else}

10.~~~~~~~~~~~~~~~~with probability $p$, $a\leftarrow$ an atom $x$ with
     minimum $\bc(x)$;

11.~~~~~~~~~~~~~~~~with probability $1-p$, $a\leftarrow$ a randomly chosen
     atom in $C$;

12.~~~~~~~~~~~~{\bf End If}

13.~~~~~~~~~~~~$\sigma\leftarrow Flip(T,\sigma,a)$;

14.~~~~~~~~{\bf End for} of $j$

15.~~~~{\bf End for} of $i$

{\bf END}\vspace*{-0.075in}
\end{flushleft}
\end{algorithm}      

We note that the procedure $\flp$ may, in general, depend on the input 
theory $T$. It is not the case in $\wsat$ and other similar algorithms 
but it is so in one of the algorithms we propose in the paper. Thus, we
include $T$ as one of the arguments of the procedure $\flp$.

We also note that in the algorithm, we use several parameters that, 
in our implementations, we enter from the command line. They are 
$\mtr$, $\mfl$ and $p$. All these parameters affect the performance 
of the program. We come back to this matter later in Section \ref{exp}.

To obtain a concrete implementation of the algorithm $\gwcc$, we need 
to define $\bc(x)$ and to specify the notion of a {\em 
flip}. In this paper we follow two basic directions. In the first of 
them, we use a simple notion of a flip, that is, we always flip just 
one atom. We introduce, however, a more complex concept of the 
break-count, which we call the {\em virtual break-count}. In the second 
approach, we use a simple notion of the break-count --- the number 
of clauses that become unsatisfied --- but introduce a more complex 
concept of a flip, which we call the {\em double-flip}.


To specify our first instantiation of the algorithm $\gwcc(T)$, 
we define the break-count of an atom $x$ in $T$ as the number of 
clauses in the CNF theory $\cpilebs(T)$ that become unsatisfied after
flipping $x$. The key idea is to observe
that this number can be computed strictly on the basis of $T$, that 
is, {\em without} actually constructing the theory $\cpilebs(T)$.  
It is critical since the size of the 
theory $\cpilebs(T)$ is in general much larger than the size of $T$
(sometimes even exponentially larger). We refer to this notion of the
break-count as the {\em virtual} break-count as it is defined {\em not}
with respect to an input $\plc$ theory $T$ but with respect to a 
``virtual'' theory $\cpilebs(T)$, which we do not explicitly construct.

Further, we define the procedure $\flp(\sigma,a)$ (it does not depend
on $T$ hence, we dropped $T$ from the notation) so that, given a 
truth assignment $\sigma$ and an atom $a$, it returns the truth 
assignment $\sigma'$ obtained from $\sigma$ by setting $\sigma'(a)$ 
to the dual value of $\sigma(a)$ and by keeping all other truth values 
in $\sigma$ unchanged (this is the basic notion of the flip that is 
used in many local-search algorithms, in particular in {\em WSAT}). 
We call the resulting version of the the algorithm $\gwcc(T)$, the {\em 
virtual break-count $\wcc$} and denote it by $\vwcc$.


The second instantiation of the algorithm $\gwcc$ that we will discuss
applies only to $\plc$ theories of some special syntactic form. 
A $\plc$ theory $T$ is {\em simple}, if $T = T^{cc}\cup T^{\mathit{cnf}}$, 
where $T^{cc}\cap T^{\mathit{cnf}}=\emptyset$ and
\begin{enumerate}
\item $T^{cc}$ consists of {\em unit} clauses $C_i= k_i X_i m_i$,
$1\leq i\leq p$, such that sets $X_i$ are pairwise disjoint
\item $T^{\mathit{cnf}}$ consists of propositional clauses
\item for every $i$, $1\leq i\leq p$, $k_i < |X_i|$ and $m_i > 0$.
\end{enumerate}
Condition (3) is not particularly restrictive. In particular, it excludes 
c-atoms $k X m$ such that $k > |X|$, which are trivially false and can
be simplified away from the theory, as well as those for which $k=|X|$,
which forces all atoms in $X$ to be true and again implies
straightforward simplifications. The effect of the restriction $m > 0$
is similar; it eliminates c-atoms with $m=0$, for which it must be that
all atoms in $X$ be false. We note that $\plc$ theories we proposed as
encodings of the graph-coloring and vertex-cover problems are simple;
the theory encoding the latin-square problem is not.

In this section, we consider only simple $\plc$ theories. Let us
assume that we designed the procedure $\flp(T,\sigma,a)$ so that it has
the following property:
\begin{description}
\item{(DF)} if a truth assignment $\sigma$ is a model of $T^{cc}$ then 
$\sigma'= \flp(T,\sigma,a)$ is also a model of $T^{cc}$.
\end{description}
Let us consider a try starting with a truth assignment $\sigma$ that
satisfies all clauses in $T^{cc}$. If our procedure $\flp$ satisfies
the property (DF), then all truth assignments that we will generate in
this try satisfy all clauses in $T^{cc}$. It follows that the only 
clauses that can become unsatisfied during the try are the propositional 
clauses in $T^{\mathit{cnf}}$. Consequently, in order to compute the 
break-count 
of an atom, we only need to consider the CNF theory $T^{\mathit{cnf}}$ 
and count how many clauses in $T^{\mathit{cnf}}$ become unsatisfiable 
when we perform a flip.

Since all c-atoms in $T^{cc}$ are pairwise disjoint, it is easy to generate
random truth assignments that satisfy all these constraints. Thus,
it is easy to generate a random starting truth assignment for a
try. Moreover, it is also quite straightforward to design a procedure
$\flp$ so that it satisfies property (DF). We will outline one such
procedure now and provide for it a detailed pseudo-code description.

\begin{algorithm}[t]
\caption{Algorithm $Flip(T,\sigma,a)$}
\begin{flushleft}
\ \\
{\bf INPUT}:~~~~~~$T$ - a simple $\plc$ theory ($T=T^{cc}\cup T^{cnf}$)\\

~~~~~~~~~~~~~~~~~~$\sigma$ - current truth assignment

~~~~~~~~~~~~~~~~~~$a$ - an atom chosen to flip

{\bf OUTPUT}:~~~$\sigma$ - updated $\sigma$ after $a$ is flipped

{\bf BEGIN}

1. ~~~~{\bf If} $a$ occurs in a clause in $T^{cc}$ {\bf and}
flipping $a$ will break it {\bf then}

2. ~~~~~~~~pick the best opposite atom, say $b$, in that clause w.r.t.
break-count;

3. ~~~~~~~~$\sigma(b) \leftarrow$ dual of $\sigma(b)$;

4. ~~~~{\bf End if}

5. ~~~~$\sigma(a) \leftarrow$ dual of $\sigma(a)$;

6. ~~~~{\bf return} $\sigma$;

{\bf END}\vspace*{-0.075in}
\end{flushleft}
\end{algorithm}  

Let us assume that $\sigma$ is a truth assignment that satisfies all
clauses in $T^{cc}$ and that we selected an atom $a$ as the third
argument for the procedure $\flp$. If flipping the value of $a$ does
not violate any unit clause in $T^{cc}$, the procedure
$\flp(T,\sigma,a)$ returns the truth assignment obtained from $\sigma$
by flipping the value of $a$. Otherwise, since the c-atoms forming the
clauses in $T^{cc}$ are pairwise disjoint, there is exactly one clause
in
$T^{cc}$, say $k X m$, that becomes unsatisfied when the value of $a$
is flipped. In this case, clearly, $a\in X$.

We proceed now as follows. We find in $X$ another atom, say $b$,
whose truth value is opposite to that of $a$, and flip both $a$ and
$b$. That is, $\flp(T,\sigma,a)$ returns the truth assignment obtained
from $\sigma$ by flipping the values assigned to $a$ and $b$ to their
duals. Clearly, by performing this {\em double flip} we maintain
the property that all clauses in $T^{cc}$ are still satisfied. Indeed
all clauses in $T^{cc}$ other than $k X m$ are not affected by the
flips (these clauses contain neither $a$ nor $b$) and $k X m$ is
satisfied because flipping $a$ and $b$ simply switches their truth
values and, therefore, does not change the number of atoms in $X$ that
are true.

The only question is whether such an atom $b$ can be found. The answer
is indeed positive. If $\sigma(a)=\tr$ and flipping $a$ breaks clause
$k X m$, we must have that the number of atoms that are true in $X$ is
equal to $k$. Since $|X| >k$, there is an atom in $X$ that is false.
The reasoning in the case when $\sigma(a)=\fa$ is similar.

A pseudo-code for the procedure is given in Figure 2.

\section{Experiments, Results and Discussion}
\label{exp}

We performed experimental studies of the effectiveness of our
local-search algorithms in solving several difficult search problems. 
For the experiments we selected the graph-coloring problem, the 
vertex-cover problem and the latin-square problem. We discussed 
these problems in Section \ref{s-plc} and described $\plc$ theories 
that encode them. To build $\plc$ theories for testing, we randomly
generate or otherwise select input instances to these search problems 
and instantiate the corresponding $\plc$ encodings. For the graph-coloring
and vertex-cover problems we obtain simple $\plc$ theories and so 
all methods we discussed apply. The theories we obtain from the 
latin-square problem are not simple. Consequently, the algorithm 
$\dwcc$ does not apply but all other methods do.

Our primary goal is to demonstrate that our algorithms $\vwcc$ and
$\dwcc$ can compute models of {\em large} $\plc$ theories and,
consequently, are effective tools for solving search problems. To 
this end, we study the performance of these algorithms and compare it 
to the performance of methods that employ SAT solvers, specifically 
$\wsat$ and $\zch$ \cite{mmz01}. We chose $\wsat$ since it is a local-search 
algorithm, as are $\vwcc$ and $\dwcc$. We chose $\zch$ since it is 
one of the most advanced {\em complete} methods. In order to use SAT 
solvers to compute models of $\plc$ theories, we executed them on the 
CNF theories produced by procedures $\cpilebc$ and $\cpilebs$ (Section 
\ref{sat}). We selected the method $\cpilebc$ as it results in most
concise translations\footnote{Our experiments with the translation 
$\cpileuc$ show that it performs worse. We believe it is due to larger 
size of theories it creates. We do not report these results here due to
space limitations.}. We selected the method $\cpilebs$ as it is
arguably the most straightforward translation and it does not require
auxiliary atoms.

For all local-search algorithms, including $\wsat$, we used the same
values of $\mtr$ and $\mfl$: 100 and 100000, respectively. The
performance of local-search algorithms depends to a large degree on
the on the value of the parameter $p$ (noise). For each method and 
for each theory, we ran experiments to determine the value of $p$,
for which the performance was the best. All results we report here
come from the best runs for each local-search method.

To assess the performance of solvers on families of test 
theories, we use the following measures.
\begin{enumerate}
\item The average running time over all instances in a family
\item The success rate of a method: the ratio of the number of theories 
in a family, for which the method finds a solution, to the total number 
of instances in the family for which we were able to find a solution
using {\em any} of the methods we tested (for all methods we set a limit 
of 2 hours of CPU time/instance).
\end{enumerate}
The success rate is an important measure of the effectiveness of
local-search techniques. It is not only important that they run fast
but also that they are likely to find models when models exist.

We will now present and discuss the results of our experiments. We 
start with the coloring problem. We generated for testing five 
families $C_1, \ldots, C_5$, each consisting of 50 random graphs with 
1000 vertices and 3850, 3860, 3870 3880 and 3890 edges, respectively. 
The problem was to find for these graphs a coloring with 4 colors 
(each of these graphs has a 4-coloring). We show the results in Table 
\ref{tab1}. Columns $\vwcc$, $\dwcc$ show the performance results for 
our local-search algorithms run on $\plc$ theories encoding the
4-colorability problem on the graphs in the families $C_i$, $1\leq
i\leq 5$. Columns $\wsat$-{\em bc} and $\zch$-{\em bc} show the performance of 
the algorithms $\wsat$ and $\zch$ on CNF theories obtained from the 
$\plc$-theories by the procedure $\cpilebc$. Columns $\wsat$-{\em
basic} and 
$\zch$-{\em
basic} show the performance of the algorithms $\wsat$ and $\zch$ 
on CNF theories produced by the procedure $\cpilebs$ (since the bounds 
in c-atoms in the case of 4-coloring are equal to 1, there is no 
dramatic increase in the size when using the procedure $\cpilebs$).
The first number in each entry is the average running time in seconds, 
the second number --- the percentage success rate. The results for local-search algorithms 
were obtained with the value of noise $p=0.4$ (we found this value to
work well for all the methods).

\vspace*{-0.12in}
\begin{table}[h]
{\scriptsize
\caption{Graph-coloring problem}
\begin{center}
\begin{tabular}{|l|r|r|r|r|r|r|r|}
\hline
Family & $\vwcc$ & $\dwcc$ & $\wsat$-{\em bc} & $\zch$-{\em bc} & $\wsat$-{\em
basic} & $\zch$-{\em
basic} \\
\hline
\hline
{\em $C_1$} & 39/96\% & 97/100\% & 27/0\% & 68/100\% & 29/100\% & 91/100\% \\
\hline
{\em $C_2$} & 40/98\% & 100/100\% & 27/0\% & 142/100\% & 29/100\% & 128/100\% \\
\hline
{\em $C_3$} & 41/100\% & 103/100\% & 27/0\% & 233/100\% & 30/98\% & 146/100\% \\
\hline
{\em $C_4$} & 41/100\% & 104/98\% & 28/0\% & 275/100\% & 30/96\% & 216/100\% \\
\hline
{\em $C_5$} & 42/96\% & 108/98\% & 28/0\% & 478/100\% & 30/96\% & 594/100\% \\
\hline
\end{tabular}
\end{center}
}
\vspace*{-0.35in}
\label{tab1}
\end{table}

In terms of the success rate, our algorithms achieve or come very close
to perfect 100\%, and are comparable or slightly better than 
the combination of $\cpilebs$ and $\wsat$. When comparing the 
running time, our algorithms are slower but only by a constant factor. 
The algorithm $\vwcc$ is only about 0.3 times slower and the algorithm 
$\dwcc$ is about 3.5 times slower. 

Next, we note that the combination $\cpilebc$ and $\wsat$ does not
perform well at all. It fails to find a 4-coloring even for a single 
graph.
We also observe that $\zch$ performs well no matter which technique is 
used to eliminate c-atoms. It finds a 4-coloring for every graph that 
we tested. In terms of the running time there is no significant 
difference between its performance on theories obtained by $\cpilebc$ 
as opposed to $\cpilebs$. However, $\zch$ is, in general, slower than 
$\wsat$ and our local-search algorithms $\vwcc$ and $\dwcc$.

Finally, we note that our results suggest that our algorithms are less
sensitive to the choice of a value for the noise parameter $p$. In
Table \ref{tab8} we show the performance results for our two algorithms
and for the combination $\cpilebs$ and $\wsat$ on theories obtained
from the graphs in the family $C_1$ and for $p$ assuming values 0.1,
0.2, 0.3 and 0.4.

\vspace*{-0.12in}
\begin{table}[h]
{\scriptsize
\caption{Coloring: sensitivity to the value of $p$}
\begin{center}
\begin{tabular}{|l|r|r|r|}
\hline
Noise & $\vwcc$ & $\dwcc$ & $\wsat$-{\em basic} \\
\hline
\hline
{\em $p=0.1$} & 16\% & 100\% & 18\% \\
\hline
{\em $p=0.2$} & 98\% & 100\% & 90\%  \\
\hline
{\em $p=0.3$} & 100\% & 100\% & 98\% \\
\hline
{\em $p=0.4$} & 96\% & 100\% & 100\%  \\
\hline
\end{tabular}
\end{center}
}
\label{tab8}
\vspace*{-0.35in}
\end{table}

We also tested our algorithms on graph-coloring instances that 
were used in the graph-coloring competition at the CP-2002 conference.
We refer to \url{http://mat.gsia.cmu.edu/COLORING02/} for details. 
We experimented with 63 instances available there. For each of these 
graphs, we identified the smallest number of colors that is known to
suffice to color it. We then tested whether the algorithms $\vwcc$, 
$\wsat$ and $\zch$ (the latter two in combination with the procedure 
$\cpilebs$ to produce a CNF encoding) can find a coloring using that 
many colors. We found that the algorithms $\dwcc$, $\wsat$ and
$\zch$ (the latter two in combination with $\cpilebs$) were very
effective. Their success rate (the percentage of instances for which 
these methods could match the best known result) was 62\%, 56\% and
54\%, respectively. In comparison, the best among the algorithms that 
participated in the competition, the algorithm MZ, has success rate of
40\% only and the success rate of other algorithms does not exceed
30\%.

For the vertex cover problem we randomly generated 50 graphs with 
200 vertices and 400 edges. For $i=103,\ldots,107$, we constructed 
a family $VC_i$ of $\plc$ theories encoding, for graphs we generated, 
the problem of finding a vertex cover of cardinality at most $i$.
For this problem, the translation $\cpilebs$ is not practical as 
translating just a single c-atom $\{in_1,\ldots, in_{200}\} i$ requires
$200\choose i+1$ clauses and these numbers are astronomically large for 
$i=103, \ldots, 107$. The translation $\cpilebc$ also does not perform
well. Neither $\wsat$ nor $\zch$ succeed in finding a solution to even
a single instance (as always, within 2 hours of CPU time/instance).
Thus, for the vertex-cover problem, we developed yet another
CNF encoding, which we refer to as {\em ad-hoc}. This encoding worked
well with $\wsat$ but not with $\zch$. We show the results in Table
\ref{tab4}. For this problem, the value of noise $p=0.1$ worked best for 
all local-search methods.

\begin{table}[h]
{\scriptsize
\caption{Vertex-cover problem: graphs with 200 vertices and 400 edges}
\begin{center}
\begin{tabular}{|l|r|r|r|r|r|r|r|r|r|}
\hline
Family & $\vwcc$ & $\dwcc$& $\wsat$-{\em bc} & $\zch$-{\em bc} & $\wsat$-ad-hoc &
$\zch$-ad-hoc \\
\hline
\hline
$VC_{103}$ & 117/100\% & 300/100\% & 11/0\% & 7200/0\% & 1696/100\% &
7200/0\% \\
\hline
$VC_{104}$ & 86/100\% & 225/100\% & 11/0\% &  7200/0\% & 1400/100\% &
7200/0\% \\
\hline
$VC_{105}$ & 69/100\% & 178/100\% & 11/0\% &  7200/0\% & 1191/100\% &
7200/0\% \\
\hline
$VC_{106}$ & 29/100\% & 78/100\% & 11/0\% &  7200/0\% & 848/100\% &
7200/0\% \\
\hline
$VC_{107}$ & 10/100\% & 27/100\% & 11/0\% &  7200/0\% & 671/100\% &
7200/0\% \\
\hline
\end{tabular}
\end{center}
}
\label{tab4}
\vspace*{-0.35in}
\end{table}

Our algorithms perform very well. They have the best running time
(with $\vwcc$ being somewhat faster than $\dwcc$) and find solutions
for all instances for which we were able to find solutions using these
and other techniques. In terms of the success rate $\wsat$, when run on
{\em ad-hoc} translations, performed as well as our algorithms but was
several (7 to 67, depending on the method and family) times slower.

As in the case of graph coloring, our algorithms again were less
sensitive to the choice of the noise value $p$, as shown in Table
\ref{tab9} (the tests were run on the family $VC_{103}$.

\vspace*{-0.12in}
\begin{table}[h]
{\scriptsize
\caption{Vertex cover: sensitivity to the value of $p$ }
\begin{center}
\begin{tabular}{|l|r|r|r|}
\hline
Noise & $\vwcc$ & $\dwcc$ & $\wsat$-{\em ad-hoc} \\
\hline
\hline
{\em $p=0.1$} & 100\% & 100\% & 100\% \\
\hline
{\em $p=0.2$} & 100\% & 100\% & 100\% \\
\hline
{\em $p=0.3$} & 100\% & 100\% & 58\% \\
\hline
{\em $p=0.4$} & 100\% & 100\% & 33\%  \\
\hline
\end{tabular}
\end{center}
}
\label{tab9}
\vspace*{-0.35in}
\end{table}

We also experimented with the vertex-cover problem for graphs of an order
of magnitude larger. We randomly generated 50 graphs, each with 2000 
vertices and 4000 edges. For these graphs we constructed a family 
$VC_{1035}$ consisting of 50 $\plc$ theories, each encoding the problem 
of finding a vertex-cover of cardinality at most 1035 in the 
corresponding graph. With graphs of this size, all compilation methods
produce large and complex CNF theories on which both $\wsat$ and $\zch$ fail
to find even a single solution. Due to the use of c-atoms, the $\plc$ 
theories are relatively small. Each consists of 2000 atoms and 4001 
clauses and has a total of about 10,000 atom occurrences. Our algorithms 
$\vwcc$ and $\dwcc$ run on each of the theories in under an hour and 
the algorithm $\dwcc$ finds a vertex cover of cardinality at most 1035 
for 9 of them. The algorithm $\vwcc$ is about two times faster but has
worse success rate: finds solutions only in 7 instances.

The last test concerned the latin-square problem. We assumed $n=30$ and
randomly generated 50 instances of the problem, each specifying values
for some 10 entries in the array. Out of these instances we constructed
a family $LS$ of the corresponding $\plc$ theories. Since these $\plc$ 
theories are not simple, we did not test the algorithm $\dwcc$ here.
The results are shown in Table \ref{tab5}. For the local-search
methods, we used the value of noise $p=0.1$.

\vspace*{-0.12in}
\begin{table}[h]
{\scriptsize
\caption{Open latin-square problem}
\begin{center}
\begin{tabular}{|l|r|r|r|r|r|r|r|r|}
\hline
$\vwcc$ & $\wsat$-{\em bc} & $\zch$-{\em bc} & $\wsat$-{\em
basic} & $\zch$-{\em
basic} \\
\hline
\hline
43/100\% & 0/0\% & 5/100\% & 250/84\% & 637/96\% \\
\hline
\end{tabular}
\end{center}
}
\label{tab5}
\end{table}                       

These results show that our algorithms are faster than the combination
of $\wsat$ and $\cpilebs$ ($\cpilebc$ again does not work well with
$\wsat$) and have a better success rate. The fastest in this case 
is, however, the combination of $\zch$ and $\cpilebc$. The combination
of $\zch$ and $\cpilebs$ works worse and it is also slower 
than our algorithms.

\vspace*{-0.1in}
\section{Conclusions}

Overall, our local-search algorithms $\vwcc$ and $\dwcc$, designed 
explicitly for $\plc$ theories, perform very well. 

It is especially 
true in the presence of cardinality constraints with large bounds 
where the ability to handle such constraints directly, without the 
need to encode them as CNF theories, is essential. It makes it 
possible for our algorithms to handle large instances of search 
problems that contain such constraints. We considered one problem in
this category, the vertex-cover problem, and demonstrated superior 
performance of our search algorithms over other techniques. For large 
instances (we considered graphs with 2000 vertices and 4000 edges and 
searched for vertex covers of cardinality 1035) SAT solvers are 
rendered ineffective by the size of CNF encodings and their complexity. 
Even for instances of much smaller size (search for vertex covers of 
103-107 elements in graphs with 200 vertices and 400 edges), our 
algorithms are many times faster and have a better success rate than 
$\wsat$ ($\zch$ is still ineffective). 

Also for $\plc$ theories that contain only c-atoms of the form $1 X 1$,
$X 1$ and $1 X$, the ability to handle such constraints directly seems
to be an advantage and leads to good performance, especially in terms 
of the success rate. In the graph-coloring and latin-square problems 
our algorithms consistently had comparable or higher success rate 
than methods employing SAT solvers. In terms of time our methods
are certainly competitive. For the coloring problem, they were slower 
than the method based on $\wsat$ and $\cpilebs$ but faster than all
other methods. For the latin-square problem, they were slower than the
combination of $\zch$ and $\cpilebc$ but again faster than other methods.

Finally, we note that our methods seem to be easier to use and more
robust. SAT-based method have a disadvantage that their performance 
strongly depends on the selection of the method to compile away
c-atoms and no method we studied is consistently better than others. 
The problem of selecting the right way to compile c-atoms away does not
appear in the context of our algorithms. Further, the performance of 
local-search methods, especially the success rate, highly depends on 
the value of the noise parameter $p$. Our results show that our 
algorithms are less sensitive to changes in $p$ than those that employ 
$\wsat$, which makes the task of selecting the value for $p$ for our 
algorithms easier.  

These results provide further support to a growing trend in
satisfiability research to extend the syntax of propositional logic by 
constructs to model high-level constraints, and to design solvers that 
can handle this expanded syntax directly. In the 
expanded syntax, we obtain more concise representations of search 
problems. Moreover, these representations are more directly aligned 
with the inherent structure of the problem. Both factors, we believe, 
will lead to faster, more effective solvers.

In this paper, we focused on the logic $\plc$, an extension of 
propositional logic with c-atoms, that is, direct means to encode 
cardinality constraints. The specific contribution of the paper are 
two local-search algorithms $\vwcc$ and $\dwcc$, tailored to the 
syntax of the the logic $\plc$. These algorithms rely on two ideas. 
The first of them is to regard a $\plc$ theory as a compact
encoding of a CNF theory modeling the same problem. One can now 
design local-search algorithms so that they work with a $\plc$ theory 
but proceed as propositional SAT solvers would when run on the 
corresponding propositional encoding. We selected the procedure 
$\cpilebs$ to establish the correspondence between $\plc$ theories 
and CNF encodings, as it does not require any new propositional 
variables and makes it easy to simulate propositional local-search 
solvers. We selected a particular propositional local-search 
method, $\wsat$, one of the best-performing local-search 
algorithms. Many other choices are possible. Whether they lead to 
more effective solvers is an open research problem.

The second idea is to change the notion of a flip. We applied it
designing the algorithm $\dwcc$ for the class of simple $\plc$ 
theories. However, this method applies whenever a $\plc$ theory $T$ 
can be partitioned into two parts $T_1$ and $T_2$ so that (1) it 
is easy to generate random truth assignments satisfying constraints 
in $T_1$, and (2) there is a notion of a flip 
that preserves satisfaction of constraints in 
the first part and allows one, in a sequence of such flips, to reach 
any point in the search space of truth assignments satisfying 
constraints in $T_1$. Identifying specific syntactic classes of 
$\plc$ theories and the corresponding notions of a flip is also
a promising research direction. 

In our experiments we designed compilation techniques to allow us to
use SAT solvers in searching for models of $\plc$ theories. In general,
approaches that rely on counting do not work well with $\wsat$, as they
introduce too much structure into the theory. The translation
$\cpilebs$ is the best match for $\wsat$ (whenever it does not lead to
astronomically large theories). All methods seem to work well with
$\zch$ at least in some of the cases we studied but none worked well for 
the vertex-cover problem. To design better techniques to eliminate 
c-atoms and to make the process of selecting an effective translation 
systematic rather than ad hoc is another interesting research direction. 

Our work is related to \cite{wal97} and \cite{pre02}, which
describe local-search solvers for theories in propositional logic
extended by pseudo-boolean constraints. However, the classes of
formulas accepted by these two solvers and by ours are different. We 
use cardinality atoms as generalized ``atomic'' components of 
clauses while pseudoboolean constraints have to form unit 
clauses. On the other hand, pseudoboolean constraints are more general
than cardinality atoms. At present, we are comparing the performance of 
all the solvers on the class of theories that are accepted by all 
solvers (which includes all theories considered here).

\section*{Acknowledgments}
\vspace*{-0.1in}

The authors are grateful to the reviewers for comments and pointers 
to papers on extensions of propositional logic by pseudoboolean 
constraints. This research was supported by the National Science 
Foundation under Grant No.  0097278.


\begin{thebibliography}{10}

\bibitem{arms02}
F.A. Aloul, A.~Ramani, I.~Markov, and K.~Sakallah.
\newblock Pbs: a backtrack-search pseudo-boolean solver and optimizer.
\newblock In {\em {Proceedings of the Fifth International Symposium on Theory
  and Applications of Satisfiability}}, pages 346 -- 353, 2002.

\bibitem{bar95}
P.~Barth.
\newblock A davis-putnam based elimination algorithm for linear pseudo-boolean
  optimization.
\newblock Technical report, Max-Planck-Institut {f\"ur} Informatik, 1995.
\newblock MPI-I-95-2-003.

\bibitem{bss94}
B.~Benhamou, L.~Sais, , and P.~Siegel.
\newblock Two proof procedures for a cardinality based language in
  propositional calculus.
\newblock In {\em Proceedings of STACS-94}, pages 71--82. 1994.

\bibitem{dg02}
H.E. Dixon and M.L. Ginsberg.
\newblock Inference methods for a pseudo-boolean satisfiability solver.
\newblock In {\em The 18th National Conference on Artificial Intelligence
  (AAAI-2002)}, 2002.

\bibitem{et01a}
D.~East and M.~Truszczy{\'n}ski.
\newblock Propositional satisfiability in answer-set programming.
\newblock In {\em Proceedings of Joint German/Austrian Conference on Artificial
  Intelligence, KI'2001}, volume 2174, pages 138--153. Lecture Notes in
  Artificial Intelligence, Springer Verlag, 2001.
\newblock Full version submitted for publication (available at
  \url{http://xxx.lanl.gov/abs/cs.LO/0211033}).

\bibitem{gn02}
E.~Goldberg and Y.~Novikov.
\newblock Berkmin: a fast and robust sat-solver.
\newblock In {\em DATE-2002}, pages 142--149. 2002.

\bibitem{li00}
C.M. Li.
\newblock Integrating equivalency reasoning into davis-putnam procedure.
\newblock In {\em Proccedings of the Seventeenth National Conference on
  Artificial Intelligence (AAAI-2000)}, pages 291--296, 2000.

\bibitem{la97}
C.M. Li and M.~Anbulagan.
\newblock Look-ahead versus look-back for satisfiability problems.
\newblock In {\em Proceedings of the Third International Conference on
  Principles and Practice of Constraint Programming}, 1997.

\bibitem{ss99}
J.P. Marques-Silva and K.A. Sakallah.
\newblock {GRASP}: {A} new search algorithm for satisfiability.
\newblock {\em {IEEE Transactions on Computers}}, 48:506--521, 1999.

\bibitem{mmz01}
M.~Moskewicz, C.~Madigan, Y.~Zhao, L.~Zhang, and S.~Malik.
\newblock Chaff: engineering an efficient {SAT} solver.
\newblock In {\em Proceedings of the Design Automation Conference (DAC)}, 2001.

\bibitem{pre02}
S.D. Prestwich.
\newblock Randomised backtracking for linear pseudo-boolean constraint
  problems.
\newblock In {\em Proceedings of the 4th International Workshop on Integration
  of AI and OR techniques in Constraint Programming for Combinatorial
  Optimisation Problems, CPAIOR-02}, pages 7--20, 2002.

\bibitem{skc94}
B.~Selman, H.A. Kautz, and B.~Cohen.
\newblock Noise strategies for improving local search.
\newblock In {\em Proceedings of the Twelfth National Conference on Artificial
  Intelligence (AAAI-94)}, Seattle, USA, 1994.

\bibitem{wal97}
J.P. Walser.
\newblock Solving linear pseudo-boolean constraints with local search.
\newblock In {\em Proceedings of the 11th Conference on Artificial
  Intelligence, AAAI-97}, pages 269--274. AAAI Press, 1997.

\bibitem{zh97}
H.~Zhang.
\newblock {SATO}: an efficient propositional prover.
\newblock In {\em Proceedings of the International Conference on Automated
  Deduction {(CADE-97)}}, pages 308--312, 1997.
\newblock {L}ecture {N}otes in {A}rtificial {I}ntelligence, 1104.

\end{thebibliography}

\end{document}